\documentclass[final,2p,times,12pt]{elsarticle}

\usepackage{graphicx}

\usepackage{float}
\usepackage{lineno}
\usepackage{xcolor}
\usepackage{amssymb}

\usepackage[margin=2.8cm]{geometry}

\usepackage{multirow}
\usepackage{booktabs}
\usepackage{changepage}
\usepackage{array}
\journal{Annals of the American Association of Geographers}

\begin{document}

\bibliographystyle{apalike}\biboptions{authoryear}

\begin{frontmatter}


\title{Urban Visual Intelligence: Studying Cities with AI and Street-level Imagery}



 \author[1,2]{Fan Zhang\corref{cor1}}
 \author[1,3]{Arianna Salazar-Miranda}
 \author[1]{Fábio Duarte}
 \author[3]{Lawrence Vale}
 \author[3]{Gary Hack}
 \author[4]{Min Chen}
 \author[5]{Yu Liu}
 \author[6]{Michael Batty}
 \author[1,3]{Carlo Ratti}

 \cortext[cor1]{Corresponding author email: cefzhang@ust.hk }
 
 \address[1]{Senseable City Lab, Massachusetts Institute of Technology, United States}
 \address[2]{Department of Civil and Environmental Engineering, The Hong Kong University of Science and Technology, Hong Kong}
 \address[3]{Department of Urban Studies and Planning, Massachusetts Institute of Technology, United States }
  \address[4]{Key Laboratory of Virtual Geographic Environment, Ministry of Education, Nanjing Normal University, Nanjing 210023, China}
 \address[5]{Institute of Remote Sensing and Geographical Information System, Peking University, China}
 \address[6]{Centre for Advanced Spatial Analysis, Faculty of the Built Environment, University College London, United Kingdom}

\begin{abstract}

The visual dimension of cities has been a fundamental subject in urban studies, since the pioneering work of scholars such as Sitte, Lynch, Arnheim and Jacobs. Several decades later, big data and artificial intelligence (AI) are revolutionizing how people move, sense, and interact with cities.   
This paper reviews the literature on the appearance and function of cities to illustrate how visual information has been used to understand them. A conceptual framework, \textit{Urban Visual Intelligence}, is introduced to systematically elaborate on how new image data sources and AI techniques are reshaping the way researchers perceive and measure cities, enabling the study of the physical environment and its interactions with the socio-economic environment at various scales. The paper argues that these new approaches enable researchers to revisit the classic urban theories and themes, and potentially help cities create environments that are more in line with human behaviors and aspirations in the digital age.



\end{abstract}


\end{frontmatter}


\section{Introduction}

Images have played a crucial and enduring role in the study of cities, providing valuable insights into the physical environment and influencing urban design theories. Throughout history, images have been essential in shaping our understanding of urban spaces, their characteristics, and their impact on individuals and communities.

From the 19th century onwards, the aesthetic value of cities was emphasized, with images serving as a medium to observe, document, and evaluate the beauty and design of urban spaces \citep{freestone2011reconciling}. Images, through their ability to capture and convey the visual qualities of cities, played a vital role in communicating and promoting normative theories that advocated for the importance of beauty in urban environments. As urban design theories evolved, the focus shifted towards understanding how people use and interact with urban spaces. In the mid-20th century, designers became concerned with the inhospitable nature of modernist urban forms and sought to develop a broader approach that could account for the performance and use of space. This epistemological shift was driven by a desire to move away from prescriptive theories about how the city ``should be'' towards an understanding of how people actually experience and utilize urban environments. Visual information obtained from images and videos became instrumental in this empirical approach, enabling researchers to assess how the physical environment influences individuals' behavior, perceptions, and social interactions \citep{jacobs2011uses,whyte1980social,appleyard1981livable, jacobs1987toward}.
One of the pioneers in using images to understand cities was Kevin Lynch, who introduced the concept of \textit{imageability} to explain the varying mental impressions individuals had of different cities. Through the use of photographs and interviews with residents, Lynch solicited, assembled, and analyzed perceptual maps to identify areas that captured citizens' attention and left a lasting impression. 
Traditional data collection methods using visual information, such as images, videos, and direct observation, offer valuable insights into the relationship between human activity and city form. However, these methods are labor-intensive and time-consuming, limiting their scalability over large spatial regions or extended time periods. Fortunately, advancements in sensing technologies and the availability of geotagged imagery data now enable more detailed and extensive examinations of cities. These developments facilitate comparisons across regions and over time, providing new opportunities for comprehensive analysis.
Despite the abundance of visual data, there is still a lack of clarity regarding the \textit{characteristics of different types of visual data} and \textit{how to derive visual information in a standardized manner}. Despite numerous studies analyzing neighborhood appearance using visual data, it remains unclear \textit{how to conceptually quantify the physical environment of a place} and \textit{how such quantification can contribute to a systematic understanding of the human-place relationship and inform classical theories and practices}.

In this paper, we examine theories and recent empirical studies on the use of visual information for understanding cities. We propose a conceptual framework called ``Urban Visual Intelligence'' that demonstrates the integration of images and Artificial Intelligence (AI) to observe, measure, and represent the physical environment's characteristics and its interaction with the socioeconomic environment.


\section{Historical overview of visual information in urban studies}
\label{sec:review}

The tradition of incorporating visual information runs through the history of modern urban studies. 
Since the early days of modern city planning, planners have documented and measured physical environmental attributes, which could be extracted from photographs or sketches taken along the streets and sidewalks. These attributes include shape, proportion, rhythm, scale, complexity, color, order, elements, and hierarchy, through which planners describe the urban forms \citep{wohlwill1976environmental}. This tradition of utilizing the formal attributes of the physical environment to create a pleasing sensory experience for citizens can be traced back to Camilo Sitte. Sitte advocated for the interpretation of cities through visual art and architecture, expressing strong criticism towards rigid symmetry and emphasizing the value of irregularity in urban form. He proposed that the aesthetic aspect of cities should be a primary consideration in their design \citep{sitte1889city}.  This focus on the design of the physical environment as a means to influence citizen behavior reached its culmination in planning utopias such as the Garden City and the City Beautiful Movement. Supporters of these philosophies believe that the beauty, order, and cleanliness of the public realm have the power to shape civic spirit and enhance the quality of life \citep{talen2002beyond}. Designers and theorists, including Frederick Law Olmsted Sr., Phillip Mackintosh, and F. W. Fitzpatrick, argued that the creation of visually pleasing cities would contribute to citizens' satisfaction, comfort, and pride. They recognized the aesthetic experience of urban spaces as a fundamental factor in urban design and planning \citep{wilson1990city, mackintosh2005development, mulford1899plate, nasar1994urban, ahlfeldt2012valuing}. However, despite a consensus that cities should be aesthetically pleasing and beautiful, the debate regarding what defines beauty in a city or a space persists. Is beauty in the eye of the beholder? Alternatively, could aesthetics be measured so that designers could apply that measure to design spaces that appeal to many? 

With these lingering questions, the focus of planners in the 20th century began to shift from the formal attributes of the physical environment, specifically aesthetics, to the subjective experiences they evoke. At its core, this approach to urban design aimed to gain a deeper understanding of how humans perceive and evaluate urban scenes visually. 
The supporting studies underscored the significance of comprehending cities through people's visual perceptions \citep{arnheim1965art}. In parallel, researchers attempted to capture how a city's physical environment can elicit emotions that help inform our understanding of attractive and unattractive environments. For instance, \citet{nasar1998evaluative}, proposed a model explaining how aesthetic responses emerge from human interaction with the surrounding environment. Similarly, \citet{rapoport1990meaning} identified 36 characteristics related to the size and shape of typical aesthetically pleasing urban environments. Overall, these studies on perception centered around how people shape their environment and, in turn, how the physical environment affects them. However, their impact on the theories and practices of urban studies remained limited due to the lack of approaches for quantifying and representing the physical environment on a large scale.

To quantify and represent the physical environment of a place, \citet{lynch1960image} introduced ``imageability'' as a new criterion, building upon ideas about perception but shifting the focus to human cognition rather than just aesthetics \citep{Lynch1984good}. Lynch emphasized the importance of meaning in understanding how people navigate and comprehend urban environments. His study revealed that as individuals traverse an environment, they accumulate spatial knowledge acquired through observation, which they translate into mental maps.   In his seminal work, \textit{The Image of the City}, Lynch proposed three categories to encapsulate the physical environment: Identity (distinct visual objects), Structure (recognizable patterns and relationships between objects), and Meaning (emotional values and character of a place) \citep{lynch1960image}. Initially, Lynch assessed these dimensions using traditional approaches employed by urbanists, such as sketch maps, field surveys, and interviews conducted within a limited number of neighborhoods and with a small participant pool. Similarly, Milgram (1970) proposed the creation of collective maps of New York City to measure its recognizability through a series of small-scale experiments \citep{milgram1970experience}.

These works laid the foundation for a shift in philosophy towards people-centered and place-based urban design in the late 20th century. Designers and planners began to embrace an emphasis on the performance, vitality, and usage of spaces as an alternative measure of urban design quality \citep{gehl1971life}. Typically, scholars collected information on how people utilize urban spaces using simple recording techniques involving pen and paper, complemented by photographic images. A classic example of this approach is William H. Whyte's influential study on the social life of public spaces, known as ``The Street Life Project,'' which employed conversations, photographs, and careful video analysis to observe people's usage of public spaces \citep{whyte1980social}. Similarly, Gehl (1971) extensively observed urban spaces to document the elements that foster liveliness and contribute to social interactions in his influential book \textit{Life Between Buildings} \citep{gehl1971life}. Overall, scholars studying human-centered urban design utilized observation and video recordings to measure human behavior and the appropriation of public areas \citep{Pushkarev1976urban}. These studies have significantly influenced urban design practices in the 21st century, with their methodologies becoming standard practices for documenting and understanding the interactions between the physical and socioeconomic environments.

While these pioneers offered groundbreaking implications for urban studies and design, it is important to acknowledge that we now live in a rapidly changing world, necessitating the repetition of these studies in different contexts and time periods to address our current questions. Additionally, contemporary researchers raise concerns about the small sample sizes and subject selection biases in the aforementioned studies, which make them susceptible to variations in preferences across different populations and over time \citep{nasar1998evaluative}.

\section{Framework of Urban Visual Intelligence}
Today, hybrid sensing techniques such as crowdsensing or ad hoc sensor deployment offer researchers diverse data to analyze city life. Additionally, urban big data and AI-driven approaches allow for the quantification of the physical environment, socioeconomic conditions, and human dynamics with unprecedented performance. With these tools, researchers can observe the interaction between human behavior and the physical environment across spatial and temporal scales. 

To explore these opportunities, we propose a framework called \textit{Urban Visual Intelligence}. This framework aims to review the data and methods that researchers adopt today and how they differ from historical methods. In particular, the framework elaborates on how visual intelligence technologies are used to observe, measure, and represent the urban physical environment and its interaction with socioeconomic environments. Importantly, this framework centers around the use of AI-based tools to analyze street-level imagery. We hope to illustrate the key issues and complementary approaches to these studies, weave together different technologies, and think from the existing literature.

\begin{figure}[H]
\centering
\includegraphics[width=1\textwidth]{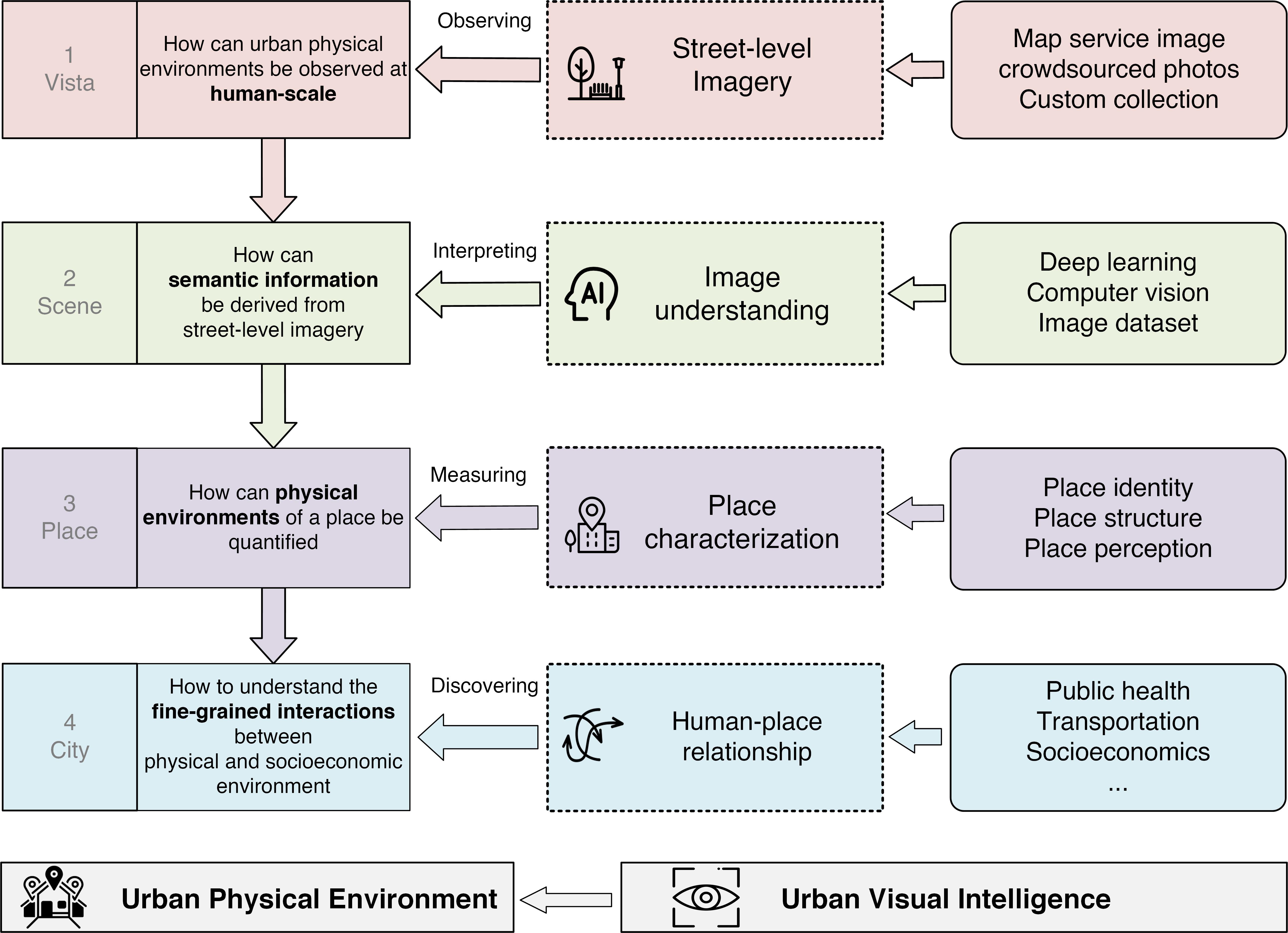}
\caption{\label{fig:uvi} Framework of Urban Visual Intelligence. 
The framework depicts the essential topics related to the urban physical environment and corresponding studies. The framework emphasizes the use of visual intelligence technologies to observe, measure, and represent physical environments, and to explore their interactions with socioeconomic dimensions at various levels and scales.
}
\end{figure}

The framework is illustrated in Figure \ref{fig:uvi}, which consists of four hierarchical levels that address four main issues. These issues include: (1) How can the urban physical environment be observed at a human scale? (2) How can semantic information be derived from street-level imagery? (3) How can the physical environment of a place be quantified? and (4) How can we understand the fine-grained interactions between the physical environment and their socioeconomic environment?

Starting from the top, the first level focuses on the data sources available to observe the physical environment, such as Google Street View and crowdsourced platforms. At this level, studies use street-level imagery to observe physical environments at the human scale, where a single image can be considered a vista. The second level focuses on the interpretation of semantic information from street-level imagery. Computer vision and deep learning techniques are used to extract data about the physical environment, such as measuring tree or sky coverage in a scene depicted in a street-level image. Moving down to the third level, studies use a collection of images to create quantitative representations of a place, which can be used to characterize place structure and perception. Finally, the fourth level takes the measurements of places and physical environments a step further to study their interactions with human dynamics and socioeconomic characteristics.

These four levels are also associated with four scales of street-level imagery usage, namely vista, scene, place, and city. Each level increases in complexity and builds upon the previous level, creating a comprehensive process for applying images to understand urban environments. In the following sections, each of the four levels of the \textit{Urban Visual Intelligence} framework will be discussed in detail, along with relevant research and works in the field.





\section{How can the urban physical environment be observed at a human scale?}
    \label{sec:data}
The first level of the \textit{Urban Visual Intelligence} framework highlights street-level imagery as a crucial data source for studying urban environments. The rapid adoption of mobile internet technologies and the increasing use of web mapping services and crowdsourcing platforms have resulted in the production of geotagged images at an unprecedented rate, covering every corner of cities \citep{goodchild2007citizens}. This new data source, commonly referred to as ``street-level imagery,'' has extensive spatial coverage and has been widely used to observe large-scale urban environments \citep{ibrahim2020understanding, biljecki2021street, Duarte2021what}. In 2020 alone, 1.43 trillion photos were taken from digital cameras of mobile phones, according to Keypoint Intelligence \footnote{https://keypointintelligence.com/}.    
    
\begin{figure}[H]
    \centering
    \includegraphics[width=1\textwidth]{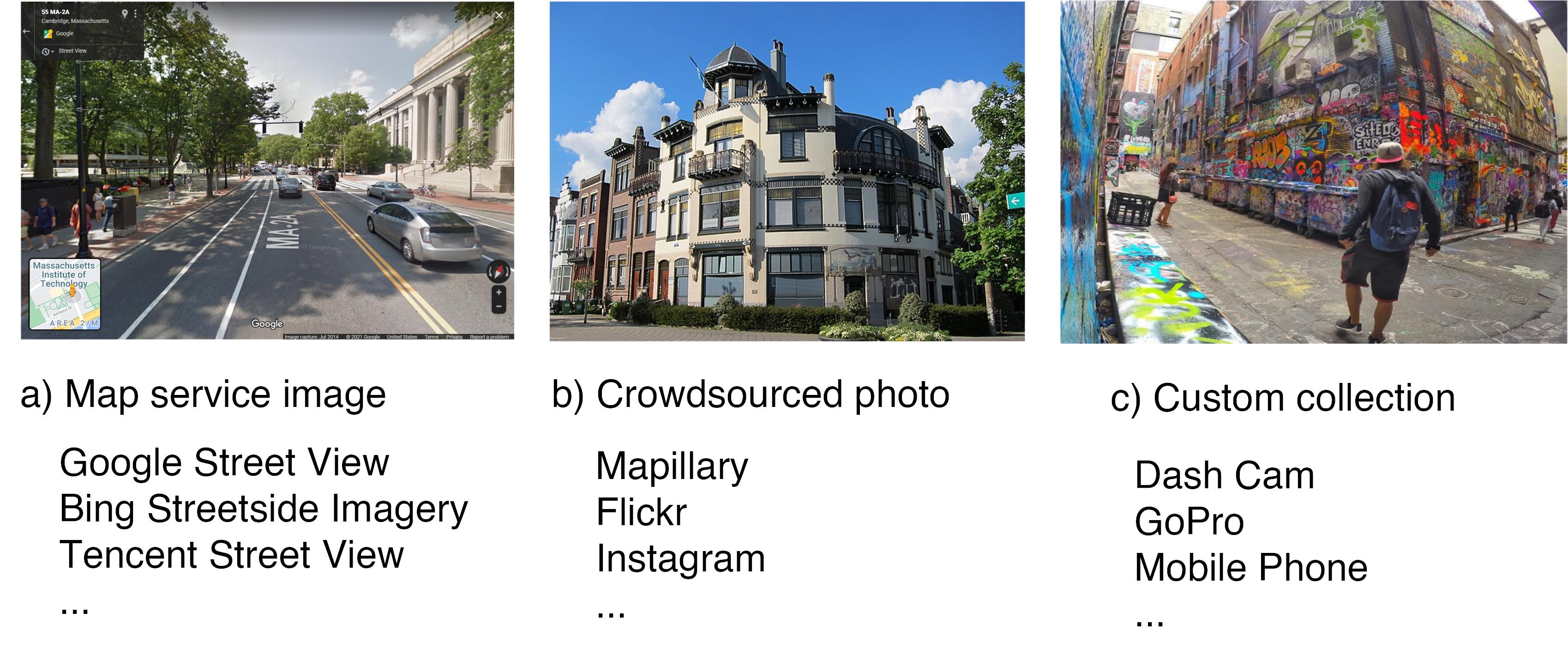}
    \caption{\label{fig:types} Three sources of street-level imagery}
    \end{figure}

Figure \ref{fig:types} illustrates three common sources of street-level imagery that can be employed for analyzing cities. The first category consists of map service images, such as Google Street View, which offer a stable update frequency, the broadest coverage (spanning over 200 countries worldwide), and a uniform standard, facilitating comparative analysis between different places \citep{anguelov2010google, goel2018estimating}. The second category comprises crowdsourced photos obtained from platforms like Flickr and Mapillary, representing a form of Volunteered Geographic Information (VGI) \citep{goodchild2007citizens}. As the volume of crowdsourced data continues to grow, with denser spatiotemporal coverage, crowdsourced photos are expected to surpass map services as the primary source of street-level imagery. Lastly, the third category encompasses custom collections of images, captured by individuals or researchers for specific research purposes. For instance, collecting a time series of images allows researchers to track changes in the physical environment and individual activities. Custom collections can complement mapping services and crowdsourced imagery.

Compared to traditional data sources used to study cities, such as personal interviews and direct observations, street-level imagery offers several advantages, including easy accessibility, extensive spatiotemporal coverage, and objective and standardized views of the physical environment from embedded vantage points \citep{rzotkiewicz2018systematic, ibrahim2020understanding, kang2020review}. Moreover, street view imagery provides a distinct perspective in comparison to satellite imagery. While satellite imagery offers an aerial view, street view imagery captures the world from a human-level perspective, which is more relatable to people's experiences and perceptions of cities. This perspective allows researchers to examine visual cues at the human scale, providing valuable insights for urban planning and design. The recent standardization of images across cities and the development of visual analytic methodologies have further facilitated the analysis of street-level imagery.
    
    
Street-level imagery has now become one of the most valuable data sources for studying physical environments \citep{biljecki2021street, cinnamon2021panoramic, he2021urban}. Its application spans various research areas, including physical environment auditing \citep{kang2018building, li2018estimating, zhang2020quantifying, ning2021exploring}, public health \citep{nguyen2018neighbourhood, keralis2020health, he2020association}, urban mobility and transportation \citep{lu2019associations, hong2020did, mooney2020development}, energy estimation \citep{liu2019towards, zhang2022quantifying, sun2022understandingEnergy}, and real estate \citep{law2019take, yang2020financial, johnson2020valuing, kang2020understanding}, among others.

\section{How can semantic information be derived from street-level imagery in a scene?}\label{sec:tech}
The second level of the \textit{Urban Visual Intelligence} framework focuses on the utilization of deep learning and computer vision techniques to derive and interpret semantic information from street-level imagery.

    \subsection{Deep learning and computer vision}

    Traditionally, visual data collected from field surveys required manual interpretation, making the process labor-intensive and limiting the scale of research. To overcome these challenges and enable large-scale studies, image processing techniques were developed to analyze visual information in bulk. However, these conventional methods were restricted to processing low-level features such as color histograms and spectral characteristics. While valuable, these approaches were unable to extract high-level information such as semantic objects, styles, and scene attributes. Understanding these features is crucial for studying the relationship between a city's physical appearance and human behavior.

    Deep learning and computer vision techniques have emerged to address these limitations and extract high-level information from images. Deep learning refers to a set of computer algorithms inspired by the neural structure of the human brain, enabling models to mimic human cognitive functions such as understanding, learning, planning, and problem-solving. Deep learning has revolutionized various fields, including speech recognition \citep{hinton2012deep}, natural language processing \citep{sutskever2014sequence}, game problem-solving \citep{silver2016mastering}, and computer vision \citep{ren2015faster,he2017mask}. These achievements are attributed to the exceptional performance of deep learning models in effectively extracting high-level information from images. In the context of urban applications, deep learning techniques offer a powerful framework for comprehending the content of urban images.

\begin{figure}[]
    \centering
    \includegraphics[width=1\textwidth]{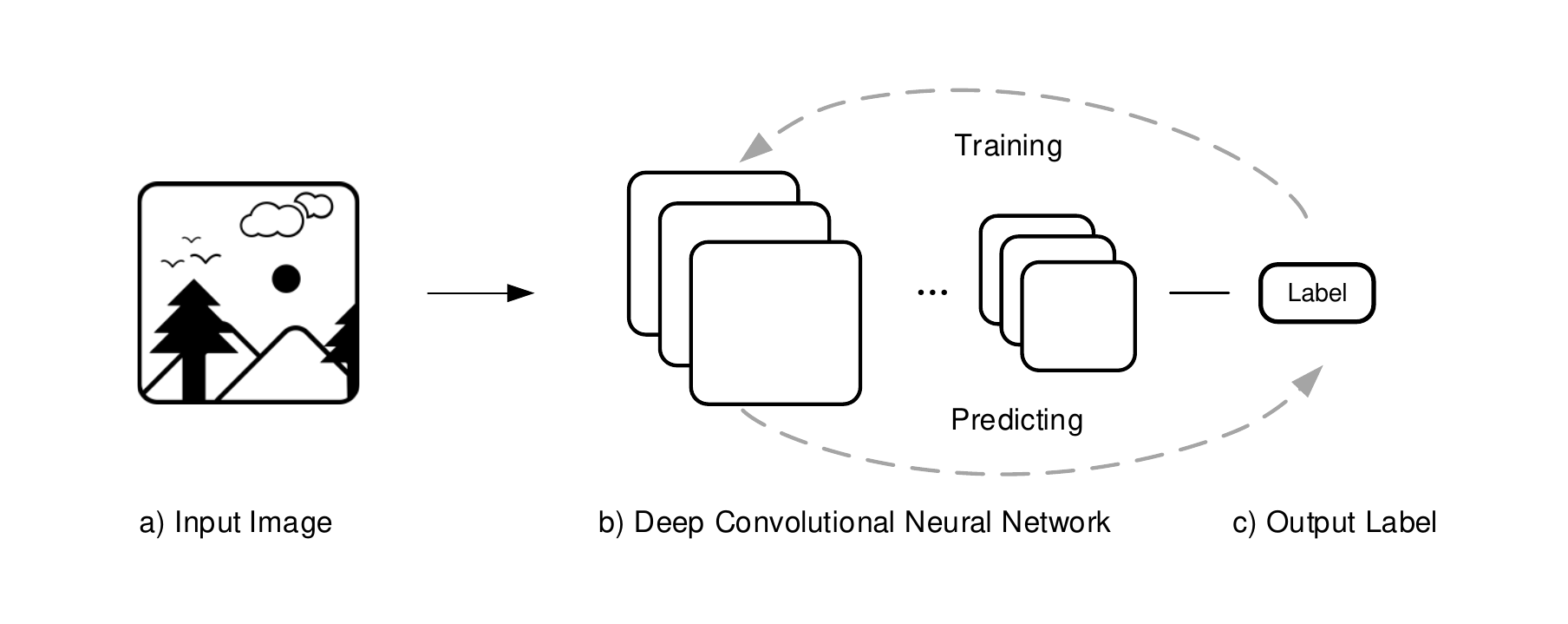}
    \caption{\label{fig:dcnn} Illustration of deep learning in urban image predictions: (a) Input urban image, (b) Deep convolutional neural network, (c) Output label.}
    \end{figure}

    Figure \ref{fig:dcnn} illustrates the functioning of a deep learning model, particularly its application in computer vision tasks such as object detection and image categorization. The core model depicted in the figure is a deep convolutional neural network (DCNN). The primary objective of a DCNN is to assign a correct label to an image, facilitating predictions about the scene or objects within the image.

    The process can be divided into two phases: training and inference. In the training phase, an image (Figure \ref{fig:dcnn}a) is inputted into a pre-designed DCNN, which processes the image layer by layer (Figure \ref{fig:dcnn}b). The final layer of the DCNN generates a predicted label, which is then compared with the true label of the image (Figure \ref{fig:dcnn}c). The parameters within the DCNN are optimized iteratively, minimizing the difference between predicted and true labels.    
    In the inference phase, the well-trained model is employed to make predictions on new images. This learning and inference process is akin to human learning, where observations are made, patterns are recognized, and feedback is used to enhance the learning process \citep{lecun2015deep}.

    \subsection{Large-scale image datasets}
 
    To train deep learning models effectively, a crucial requirement is the availability of large datasets. Deep learning models rely on a significant number of labeled images as ground truth to learn the complex relationships between inputs and their corresponding labels. Ideally, an image dataset should offer broad coverage and density. ``coverage'' implies a nearly exhaustive representation of categories, encompassing a wide range of examples. Meanwhile, ``density'' refers to having a substantial sample size of images that adequately capture the diversity within each predicted category \citep{zhou2017places}.

  Constructing a deep learning training set typically involves three main methods: labeling, matching, and synthesizing. The ``labeling'' method entails manually annotating images with categorical labels (e.g., park or parking lot) or marking object boundaries (e.g., vehicles or pedestrians) by human experts. Online tools and services have been developed to streamline this labor-intensive and time-consuming process. For instance, LabelMe is a web-based application that facilitates large-scale image annotations and online sharing \citep{russell2008labelme}. Additionally, platforms such as Amazon Mechanical Turk provide a crowdsourcing platform for on-demand image labeling tasks \citep{sorokin2008utility}. AI-assisted labeling is also gaining popularity, where pre-trained AI models assist in annotating object boundaries, even when the semantic category of the object is unknown. This approach simplifies the task for human annotators \citep{chen2020survey}. The ``matching'' method involves associating existing labels with images based on specific connections, such as co-occurrence and geographical relations. For example, to explore how the visual appearance of a house relates to its price, one can collect a large sample of houses from a real estate market website, each accompanied by a house photo and its corresponding price. Platforms such as Flickr \citep{li2013spatial} and Panoramio \citep{zhang2019discovering} offer an extensive collection of photos labeled with the cities in which they were taken, enabling city identification based on images. In estimating socioeconomic characteristics from street-level imagery, Google Street View images can be linked to demographic or human trace data using geographic coordinates (latitude and longitude) \citep{suel2019measuring, zhang2019social, ilic2019deep}. In cases where finding suitable image samples is challenging, the ``synthesizing'' method is employed. For instance, researchers have used artificial synthesis of text and street images to create datasets for training computer vision models that recognize and classify different typefaces on business signs \citep{ma2019typeface}.
   
    Prominent urban image datasets widely used in urban studies and geospatial analytics include Places2 \citep{zhou2017places} and ADE20K \citep{zhou2017scene}. The Places2 dataset consists of approximately 10 million labeled images representing various place types, such as residential neighborhoods, highways, and parks. It enables deep learning models to classify scene types using street view images as input. Similarly, the ADE20K dataset contains over 20,000 labeled images encompassing diverse visual object categories, including plants, sky, vehicles, and buildings. In addition to these datasets, researchers are compiling other data sources that connect images with ground-truth data for specific applications, such as describing scene attributes \citep{patterson2012sun}, classifying architectural styles \citep{xu2014architectural, sun2022understanding}, tracking neighborhood change \citep{Naik2017Computer}, and detecting informal settlements \citep{ibrahim2021urban}.
    
    \subsection{Existing DCNN models for urban image inference } 
     
\begin{figure}[H]
        \centering
        \includegraphics[width=1\textwidth]{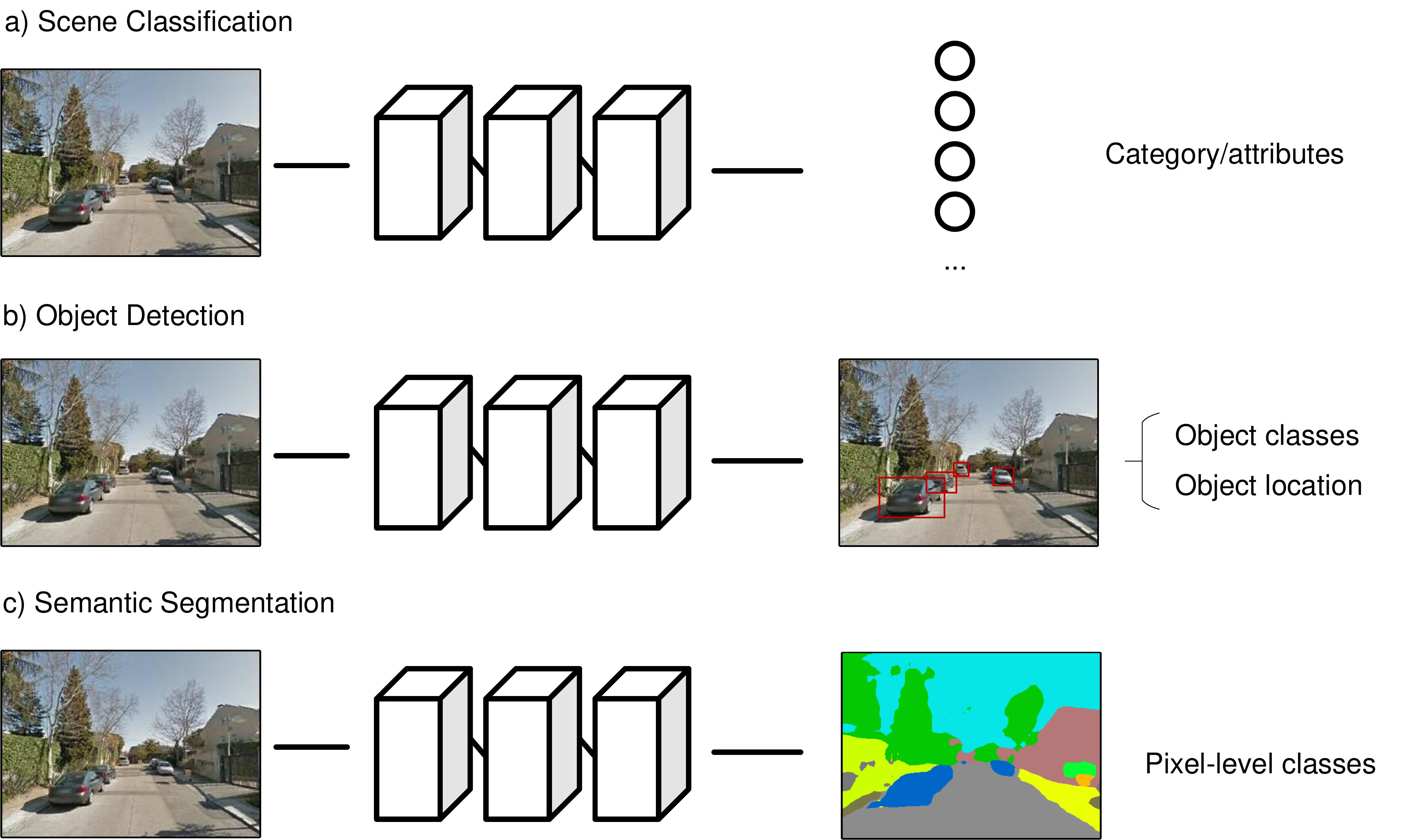}
        \caption{\label{fig:dcnn3} Three DCNN architectures in urban applications: scene classification, object detection and semantic segmentation}
    \end{figure}

   The availability of large-scale image datasets has facilitated the development of more complex deep convolutional neural network (DCNN) models with increased depth. In urban applications, DCNN models can be categorized into three main network architectures: scene classification, object detection, and semantic segmentation. Figure \ref{fig:dcnn3} illustrates the distinctions between these architectures, primarily observed in the final layers of the DCNN.
    For scene classification (Fig. \ref{fig:dcnn3}a), the last layer is a classifier that outputs a single label describing the attributes or categories of the scene in the image. Well-known DCNN architectures used for classification tasks include ResNet \citep{he2016deep}, GoogLeNet \citep{szegedy2015going}, and DenseNet \citep{huang2017densely}, among others.
    Object detection (Fig. \ref{fig:dcnn3}b) aims to identify objects within an image and provides both the predicted class of the object and the coordinates of the bounding boxes surrounding it. Popular models for object detection include Faster R-CNN \citep{ren2015faster}, SSD \citep{liu2016ssd}, and YOLO \citep{bochkovskiy2020yolov4}.
    Image segmentation (Fig. \ref{fig:dcnn3}c) involves partitioning the image into distinct segmented parts and generating pixel-wise masks for each object in the image. Widely used models for image segmentation tasks include PSPNet \citep{Zhao2016-fx}, Mask RCNN \citep{he2017mask}, and HRNet \citep{wang2020deep}, among others.
    
    \subsection{Scene understanding}
    
The extraction of scene elements is a widely employed approach for quantifying and analyzing physical environments. This can be achieved through the utilization of object detection DCNN models or object segmentation models discussed earlier. Object detection models provide the detected objects along with their bounding boxes, enabling the counting of different objects within an image. On the other hand, scene extraction models predict the object categories for each pixel, allowing further processing to calculate the object proportions within a scene. Both models offer quantitative methods for scene measurement.

An exemplary application highlighting the use of DCNN models in quantifying objective attributes of the physical environment is the \textit{Treepedia} project\footnote{http://senseable.mit.edu/treepedia}. Researchers in this project trained a deep learning model using Google Street View images to predict and classify the tree canopy of streets. By employing this scalable method instead of manual audits, they analyzed the availability of green canopy along streets in 30 cities worldwide \citep{seiferling2017green,li2018mapping}. Other related studies have combined green canopy measurements with satellite imagery-derived green indices to examine how perceptions of the physical environment are influenced by different camera angles \citep{wang2019urban,laumer2020geocoding,kumakoshi2020standardized}.

In addition to the classification of green canopy, deep learning models have been utilized to classify various elements of streets, including the sky, road, buildings, vegetation, vehicles, and pedestrians \citep{zhang2018representing,zhou2019semantic}. Access to high-quality imagery enables the classification of intricate aspects such as street signs, abandoned houses, sidewalk cracks, broken windows, and deteriorating walls \citep{less2015matching,zou2021detecting}. For example, \citet{miranda2021desirable} demonstrates the use of Google Street View to measure objective urban design characteristics that urban planners consider appealing to pedestrians\footnote{https://senseable.mit.edu/desirable-streets/}. Using Google Street View data from Boston, they calculate urban furniture, sidewalk measures, facade complexity (variation in building front materials), and visual enclosure (how well streets are defined by vertical elements like trees, walls, and buildings). Such measures hold the potential to assist urbanists in understanding which environments are more pedestrian-friendly \citep{Bettencourt2021Street}. 

    
 DCNN models offer more than just extracting physical environment elements; they are also capable of image classification. Leveraging a dataset of 10 million social media photos with detailed semantic labels, \citet{zhou2017places} trained an object segmentation model. This model effectively infers the corresponding place type (e.g., residential neighborhood, bus station, public square) and captures the key attributes describing each place (e.g., man-made, messy, sunny). Widely recognized as a benchmark, this model has been extensively utilized to gain profound insights into the functionality of various places \citep{xiao2020characterizing, zhu2020understanding, ye2020urban}.
Furthermore, street canyon classification provides a complementary example of leveraging image processing to derive valuable insights into the physical environment. Traditionally, street canyon attributes, such as building height and street width, required precise measurements using sophisticated instruments. However, \citet{hu2020classification} demonstrates how deep learning models can accurately classify street canyons using Google Street View images, eliminating the need for costly and time-consuming measurements. This approach not only reduces expenses but also enhances efficiency in analyzing different street canyon types.

Scene inference models possess the remarkable ability to extract information that extends beyond the visible content of images. These models can infer scene details that are not directly observable, such as crime rates \citep{khosla2014looking}, real estate values \citep{law2019take}, and temporal shifts in human dynamics \citep{zhang2019social}. For instance, \citet{khosla2014looking} developed a deep learning model capable of ``looking beyond the visible scene,'' enabling predictions of objects and features that may not be present in a given street view image. Surprisingly, the model can even estimate the distance to the nearest grocery store or hospital, even when these amenities are located far from the specific image. These models are trained using an end-to-end learning process, where the model autonomously learns the intricate relationships between the initial input image and the final output labels, such as crime rates and housing prices, without requiring explicit indication of the most crucial visual cues. The underlying assumption is that the built and socioeconomic environments are intimately intertwined, despite their complex and nonlinear relationship.

\section{How can the physical environment of a place be quantified?} \label{sec:rep}

While previous sections (Sections \ref{sec:data} and \ref{sec:tech}) have primarily delved into leveraging computer vision and deep learning algorithms to objectively assess the visual attributes of the physical environment, it is crucial to acknowledge the disparity between the measurements made by these algorithms and the rich tapestry of human-centered places \citep{tuan1979landscapes}.

The concept of ``place,'' deeply rooted in geography, serves as an integrating framework encompassing natural and social science perspectives \citep{patterson2005maintaining, goodchild2011formalizing}. Given its inherent complexity and subjectivity, quantifying place was historically deemed unattainable, and computational representation of a place as a whole appeared insurmountable. However, recent arguments challenge this notion of impossibility \citep{janowicz2022six}. A substantial body of research now successfully models various dimensions of place, including human activities, cognitive regions, and semantics \citep{gao2017data, purves2019places}. The availability of formal computational representations of place is essential for modern interdisciplinary research endeavors \citep{janowicz2022six}.

In this context, we capitalize on the technical advancements discussed earlier while shifting our focus to quantitatively representing and analyzing the physical environment of a place from three perspectives: place identity \& similarity, place structure, and place perception (the third level of the ``Visual Intelligence framework''). These three dimensions have been identified as pivotal in determining a place's ``imageability'' \citep{lynch1960image}. Moreover, a comprehensive characterization of places holds significant importance in geography and urban planning studies, where the integration of natural and social science concepts is vital for comprehending cities \citep{patterson2005maintaining, morison2002location}.

    \subsection{Place Identity and Similarity}
Quantitatively measuring, assessing, and understanding how humans sense places is crucial to the field of study. A place can be effectively represented by a single image or a collection thereof. The visual identity of a place pertains to its representativeness, indicating the degree of similarity or distinctiveness that allows people to easily identify it.
    
Deep learning models offer a valuable opportunity to measure the visual identity of places across various neighborhoods and cities worldwide. In practice, the measurement of visual identity and similarity can be approached as a discriminative classification problem utilizing a DCNN (Deep Convolutional Neural Network) model. Initially, the model is trained to predict the origin of a given image, i.e., the place it belongs to. Subsequently, the misclassification rates predicted by the model for each place can be employed as a metric for measuring the similarity between two places. Places that share similar inherent sample distributions are more prone to misclassification. Furthermore, the accuracy of the model in predicting a place can serve as an indicator of the distinctiveness of that place. A higher accuracy value implies that the scenes within a place are less likely to be confused with scenes from other places. Finally, the model's confidence scores for each input image can be ranked to identify scenes that possess the highest level of place representation. The confidence score signifies the model's certainty in predicting that a given scene corresponds to the specific place it was captured in. 
    
    Based on the process outlined above, a number of papers have attempted to measure place identity and similarity at different geographic scales. For instance, \citet{Doersch2012-fj} developed an automated approach to identify the distinctive architectural elements of a city that differentiate it from others. They show that visual elements, such as windows, balconies, and street signs, can distinguish Paris from other cities. On a global scale, \citet{zhang2019discovering} trained a deep learning model to recognize places among 18 cities around the world. They measured the visual similarity and distinctiveness of the cities and also identified the unique visual cues of each city (such as landmarks, historical architecture, religious sites, and unique cityscapes). For indoor spaces, \citet{zhang2016indoor} analyzed the subtle distinctions of corridors and spaces in the large interconnected buildings on the MIT campus to understand the visual elements of indoor design and human cognition that facilitate indoor navigation. Similarly, \citet{wang2019quantifying} evaluated two train stations' legibility in Paris and show how a computer vision model can identify the space from which a given photo was taken. The process through which the computer vision model identifies space is analogous to the process that pedestrians use to navigate spaces and can therefore be used to aid pedestrian routing. 
    \citet{Liu2016-yg} reproduced the \textit{Image of the City} using two million geotagged photos of 26 cities collected from a photo-sharing platform. The study yielded a series of cognitive maps of each city, demonstrating how digital techniques can revisit and enhance our understanding of places across cities. This digital approach to measuring place identity has also been extended to many other cities in recent years \citep{Salesses2013collaborative, zhou2014recognizing,filomena2019computational,  huang2021image}.


    \subsection{Place Structure}
    The street-level imagery and computer vision techniques discussed in Section \ref{sec:tech} outline the foundation to extract visual elements from images. This process can be used to further understand ``place structure.'' By place structure, we refer to an understanding of the composition and hierarchical relationships embedded in the visual elements of an image that might be important to represent a place quantitatively.

   Complementing the handful of structural elements proposed by Lynch \citep{lynch1960image}, recent papers adopt complementary perspectives to conceptually organize scene elements and scene types into categories \citep{patterson2012sun,zhou2017places,zhang2018representing}. For example, \citet{zhang2018representing}, organized hundreds of object categories that commonly appear in cities into a hierarchical tree based on their conceptual relationship. For example, ``tree,'' ``flower,'' ``grass'' are sorted into the conceptual category ``vegetation.'' The ``vegetation'' category is combined with ``waterbody'' and ``sky'' to form a broader conceptual category ``natural.'' This hierarchical semantic tree enables researchers to understand the visual structure of a neighborhood qualitatively---by understanding the presence of elements of a street at different levels and quantitatively---by measuring the abundance of scene elements using a pre-trained deep learning model. With enough images for a place, this hierarchical organization can help measure the ``structure'' of any given place.
   
    \subsection{Place Perception}

    Understanding how human perceive their surrounding environment can help assess and evaluate the quality of urban design. This topic has long been of interest to a wide variety of fields, ranging from human geography, and urban planning, to environmental psychology \citep{kaplan1989experience, lynch1960image,Tuan1977Experience, nasar1997landscapes}.
    Street-level imagery and deep learning techniques are opening up new possibilities to measure human perception. In particular, access to crowdsourced information collected online allow researchers to measure preferences and perceptions at an unprecedented scale. A key example of this approach is the online platform ``Place Pulse,'' launched to collect online ratings to evaluate human perception \citep{Salesses2013collaborative}. The project collected online volunteers' ratings on Google Street Views along six dimensions: ``safe'', ``lively'', ``beautiful,'' ``wealthy,'' ``boring'' and ``depressing.'' The platform operated for over 5 years and collected around one million ratings on 110,000 street views from more than 80,000 volunteers. Crowdsourcing platforms as such complement traditional data collection methods in multiple ways. First, the collected data represent a broad selection of people from different gender, ages, and diverse racial and cultural backgrounds. Collecting information from such a wide range of participants was inconceivable using interviews or other traditional data collection methods. Second, the evaluation of thousands of street scenes (56 cities from 28 countries worldwide) allows researchers to account for framing effects and the consistency of respondents, which small sample questionnaires cannot afford to do. 
    
    The rise of crowdsourcing platforms like Place Pulse has enabled a series of studies focused on how humans visually evaluate their surroundings \citep{ordonez2014learning,dubey2016deep}. Studies have revisited classic urban theories focused on the relationship between the physical environment and perceptions that could not be tested before due to small sample sizes and geographic scale limitations. For example, \citet{zhang2018measuring} examined the spatial distribution of human perceptions in Beijing and Shanghai using one million street views and image segmentation techniques. In particular, the study explores how street features affect human perceptions and also measures whether the physical disorder of a place (measured using litter, graffiti, and poorly maintained buildings as proxies) has a negative effect on people's feelings, providing an effective tool to evaluate the ``sense of place'' of large-scale urban areas.  \citet{saiz2018crowdsourcing} uses the ubiquitous posting of millions of photographs online to understand how people value the aesthetic dimension of the physical environment. They show that street-level imagery offers a scalable way to measure subjective attractiveness across and within cities, enabling us to build a more comprehensive understanding of how people perceive their surroundings.
    
    Human perceptions of the physical environment derived from DCNN methods have also been used to measure cities' social and economic dynamics. Research on this topic has focused on using street-level imagery obtained from Google Street View to measure changes in the neighborhood's physical appearance. \citet{Naik2017Computer} relate changes in the physical appearance of five US cities with economic and demographic data to document the underlying factors that predict neighborhood improvement. \citet{zhang2020uncovering} characterize a place in terms of physical appearance and popularity, discovering many unassuming but popular restaurants in Beijing. Locals frequently visit a host of places for social engagements despite their common location on deep alleys of old neighborhoods with unappealing appearances.
   

\section{How can we understand the fine-grained interactions between the physical environment and their socioeconomic environment?}

The fourth level of the \textit{Urban Visual Intelligence} framework delves into the study of fine-grained interactions between the built environment and the socioeconomic context. This understanding holds significant importance in disciplines such as geography, environmental science, social science, urban studies, and planning. In this section, we highlight practical applications focusing on three major topics: public health, transportation, and the socioeconomic environment of places. While these topics do not encompass the entirety of street view imagery research, they serve as prominent areas that exemplify exciting new applications.
 
\subsection{Public health}
Conventional environmental health studies have relied on field surveys and questionnaires to characterize the physical environment. Researchers and participants in such studies typically record and describe the study area's physical attributes using predetermined survey forms \citep{ball2001perceived, takano2002urban, lawlor2003geographical, gullon2015assessing}. Other studies utilize spatial analysis and GIS to derive environmental characteristics, such as employing space syntax approaches or measuring accessibility indicators \citep{pliakas2017optimising, leslie2008perceptions}. However, street-level imagery and visual intelligence offer a complementary perspective to these established methods. They facilitate cross-country comparisons as the imagery is collected from various countries and provide information captured from a human standpoint \citep{biljecki2021street}.
The physical environment exerts an impact on health outcomes through various mechanisms, encompassing physical aspects (e.g., obesity) and psychological factors (e.g., mental health) \citep{mitchell2008effect, ulrich1984view, lee2012effect, mehrabian1974approach}. Street-level imagery has proven valuable in measuring visual features that are linked to health. These features include exposure to green spaces, visual enclosure, presence and quality of sidewalks, urban infrastructure and amenities, food advertisements, and visual indicators of physical disorder. For instance, fine-scale measurement of greenery has been utilized to understand walking and cycling behaviors \citep{lu2018effect, lu2019associations}, its impact on children's body weight \citep{yang2020urban}, mental health \citep{svoray2018demonstrating, kang2019extracting, james2015review}, and perceived safety \citep{li2015does, kruse2021places}. Comparisons have also been made between greenery metrics derived from street view imagery and remote sensing imagery (e.g., normalized difference vegetation index, NDVI), highlighting the advantage of using street-level imagery to capture eye-level greenery on streets \citep{villeneuve2018comparing,lu2019associations,larkin2019evaluating}. Street view imagery and remote sensing imagery capture distinct yet complementary aspects of natural environments \citep{helbich2019using, larkin2019evaluating, kang2020review}.

Empirical studies have further revealed associations between physical aspects derived from street view imagery and health outcomes. For instance, visually enclosed streets have been found to correlate with higher quality, while the presence of sidewalks and crosswalks is linked to greater walkability and improved mental health \citep{vargo2012google,yin2016measuring, nguyen2018neighbourhood,li2018investigating,wang2019relationship}. Features extracted from street view imagery, such as food and beverage advertisements, have been used to identify obesogenic environments \citep{feuillet2016neighbourhood,roda2016mismatch, egli2019viewing}, and visual cues such as visible utility wires overhead have served as proxies.

\subsection{Transportation and mobility}

\subsection{Transportation and Mobility}
Street-level imagery provides valuable insights into transportation behavior and its consequences by utilizing features derived from the physical environment. These features, such as road infrastructure, can enhance virtual audits, enabling the identification of traffic black spots \citep{tanprasert2020recognizing} and potential urban congestion areas \citep{qin2020graph}. This section focuses on how traffic and physical environment characteristics extracted from street-level imagery contribute to understanding their association with transportation behavior.

In addition to virtual audits, studies have utilized image-derived features to examine transportation behavior. For example, researchers have found that specific road characteristics, such as traffic lights, density of speed bumps, and number of pedestrian crossings, are related to traffic volumes and route choice behavior \citep{verhoeven2018differences,den2020neighbourhood}. Other road features, including the number and width of bicycle lanes, as well as sidewalk and road surface conditions, have been used to explain variations in pedestrian crashes and traffic accidents \citep{johnson2015injury,isola2019google, hu2020investigation,kwon2020examination,mooney2020development}. These efforts contribute to better city planning and aid in testing interventions to improve pedestrian and vehicle safety. For instance, \citet{miranda2021desirable} measured pedestrians' deviations from the shortest route to construct a measure of street desirability. The study employed computer vision techniques to measure various physical environment characteristics, such as the presence of urban furniture, parks, visual enclosure, and facade heterogeneity, to identify common attributes of desirable streets. By quantifying these urban design features and relating them to pedestrian behavior, researchers can track how streets change over time, assisting practitioners in identifying areas affected by blight or perceived as hazardous, thus enabling focused efforts on revitalizing distressed streets.

Deep learning approaches offer a non-linear modeling approach to studying the associations between the physical environment and urban mobility. The appearance of the physical environment captured in images can provide insights into its function and land use type \citep{qi2011measuring,liu2012urban,yuan2012discovering,fan2021rhythm}. Deep learning models can capture the non-linear associations between these aspects through ``End-to-End training.'' For example, \citet{zhang2019social} inferred hourly human activity intensity at the street level from street view images, even in the absence of pedestrians or vehicles in the images. The results demonstrate the potential of deep convolutional neural network (DCNN) models to learn high-level street view imagery features that can explain up to 66.5\% of the hourly variation in urban mobility. Similar approaches have been applied to predict spatial patterns of cycling and walking using points of interest and street view images \citep{chen2020estimating, hankey2021predicting}.

Computer vision and deep learning approaches hold great promise for researchers seeking to understand how the physical environment can be designed to guide people's use of cities. For instance, \citet{mirowski2018learning} applied deep reinforcement learning to teach agents to navigate cities without a map. By observing street view images alone, the agent can learn physical environment features that assist in traversing distances of several kilometers to reach destinations. Strategically placing visual elements and infrastructure can aid the decision-making process for individuals as they navigate through cities.

\subsection{Socioeconomic Characteristics}
The physical environment can provide valuable insights into the socioeconomic characteristics of a city. With the support of street-level imagery and deep learning, there has been an increased interest in fine-grained characterization of the physical environment and its interactions with social and economic outcomes, including income, real estate, and crime \citep{ibrahim2020understanding,biljecki2021street}.

Crime is a prominent socioeconomic dimension that has been extensively studied using street-level imagery and deep learning techniques \citep{zhou2021using}. The motivation behind this research is the understanding that sustainable communities need to be safe from crime and also be perceived as safe by their residents. To explore the relationship between the appearance of safety and actual crime rates, \citet{zhang2021perception} propose a measure called ``perception bias,'' which quantifies the mismatch between people's perception of safety inferred from Google Street View images and the actual incidence of violent crime. This study also investigates the socioeconomic factors associated with this perception bias.

The visual quality of neighborhoods has proven to be an effective predictor of real estate values and housing appreciation \citep{yang2020financial,kang2020understanding, kang2021human,qiu2022subjective}. Elements captured in images, such as specific types of vehicles, can accurately predict neighborhood demographics and political tendencies \citep{gebru2017using}. Similarly, the presence of particular typefaces used in business amenities can serve as proxies for neighborhood income \citep{ma2019typeface}.

In addition to static physical environment features extracted from street view images, analyzing images captured at different time periods can provide valuable insights into how the physical environment is changing. For example, \citet{Naik2017Computer} created a metric of physical urban change by using images collected at different time intervals to test theories related to human capital agglomeration and the tipping point theory of urban change. The results of this study demonstrate that infrastructure improvements in neighborhoods are associated with education and population density, and that neighborhoods with better initial appearances experience more significant improvements over time.

Similar to the ``End-to-End training'' strategy in transportation, deep convolutional neural networks (DCNNs) can be trained using street-level imagery to predict socioeconomic characteristics, enabling researchers to capture the complex relationships between the physical environment and socioeconomic factors. This approach has been used to study job-housing patterns \citep{yao2021delineating}, social and environmental inequalities \citep{suel2019measuring}, and income, overcrowding, and environmental deprivation in urban areas \citep{suel2021multimodal}.

\section{Discussion}
\label{sec:dis}

\subsection{Towards Urban Visual Intelligence: what it can address and what it misses}

In this paper, we have demonstrated the utilization of images and deep learning techniques in the study of the visual aspects of cities and their connection to broader concerns regarding the performance of urban spaces. However, certain dimensions of cities cannot be comprehended solely through images. In this section, we explore the limitations of using digitally collected and processed visual information for understanding urban environments.

\textbf{Visual detection tasks.} Visual detection of elements from images constitutes a fundamental aspect of the \textit{Urban Visual Intelligence} framework. As demonstrated, advancements in deep learning models enable the classification of elements depicted in an image (vehicles, buildings, and vegetation), the human activities within it (walking, talking, and queuing), and the type of scene it represents (park, parking lot, and residential neighborhood). 
These characteristics of the physical environment can be accurately detected due to their relatively consistent nature across different geographical contexts and time periods. Consequently, the effectiveness of visual detection tasks relies on the modeling capabilities of DCNNs, which have already attained impressive capacities and continue to improve. Therefore, visual detection is unlikely to be the primary focus of future research.

\textbf{Within-place and between-place inference.} 
Within-place inference refers to the extent to which an attribute extracted from an image can predict various outcomes for a place, ranging from the scale of a single block to that of a region. In an era where the physical environment and social dimensions are deeply intertwined, numerous aspects of a city are interconnected \citep{batty2021defining}. Consequently, conventional modeling methods may overlook these intricate and nonlinear relationships. Traditional statistical models have been employed to examine the association between a specific attribute of a city (e.g., greenery density) and another attribute (e.g., neighborhood health outcomes). However, DCNN models can tackle this task with enhanced nonlinear modeling capabilities. By assigning a label to a DCNN model, it can effectively identify relationships—whether linear or nonlinear—between input street-level imagery and output variables (such as the socioeconomic composition of a neighborhood, real estate prices, or human activity density). Therefore, the performance of ``within-place inference'' depends on the strength of the underlying relationships between the physical (visual) environment and its corresponding social correlates.

``Between-place inference'' refers to the applicability of a DCNN model fitted in one place to another location. This type of inference confronts three sets of issues that can be analyzed from three complementary disciplinary fields: machine learning, urban studies, and GIScience.

From the perspective of machine learning, between-place inference commonly encounters challenges regarding cross-domain generalizability \citep{neyshabur2017exploring}. The limited generalizability of models primarily arises from differences in underlying data distributions, including relationships between input images and output labels, as well as variations between the training and testing domains. This issue, known as ``domain shift'' \citep{quinonero2009dataset}, has been extensively investigated in machine learning and can be addressed through domain adaptation techniques \citep{wang2018deep}.

Regarding urban studies, between-place inference has predominantly focused on challenges related to the measurement of a ``place.'' The heterogeneous uses of places and their perception varying across cultures make between-place inference challenging. For instance, a DCNN model trained to infer urban mobility patterns using street-view images in China may not perform well in Western cities due to significant variations in human activity patterns, even when the two cities have similar street shapes \citep{zhang2019social}. Places are heterogeneous due to differences in culture, geographical context, climate, historical development, and various other factors. Consequently, it is unlikely that an inference model developed for one place can be directly applied to another without domain adaptation.

Finally, from the perspective of GIScience, between-place inference faces issues concerning replicability in spatial analysis \citep{goodchild2020introduction,kedron2021reproducibility,goodchild2021replication}. Challenges associated with replicability stem from the spatial heterogeneity and non-stationarity inherent in spatial data. Inferences between locations are difficult to make due to spatial variation, leading to non-invariant results across different areas. Moreover, spatial non-stationarity implies that relationships between variables may not be consistent across places, making generalization across contexts problematic. Therefore, the local nature of a DCNN model often hampers its performance in different locations. Incorporating these principles into DCNN models provides an opportunity to enhance their generalizability and transferability across various contexts \citep{li2021tobler}.

\textbf{Cultural and subjective meaning.} 

A place's definition goes beyond its natural and constructed surroundings; it also encompasses the cultural and subjective significance attributed to it by people. As a cultural landscape, a place obtains a unique meaning through its inhabitants, activities, events, and historical evolution. For instance, the experience of visiting the Eiffel Tower in Paris cannot be adequately captured by merely seeing a replica of the tower in Las Vegas or Shenzhen. While the physical environment contains important social and cultural dimensions, its visual representation may not always reveal the nuanced meaning it holds for locals or visitors. A row of brick houses in Edinburgh may appear similar to its counterpart in Boston, but their meanings can differ significantly.

This understanding aligns with the concept of special geography \citep{warntz1989newton} (or idiographic science), which assumes that each place is distinct and possesses unique properties that cannot be replicated. Conversely, most machine learning models rely on an inductive learning process, aiming to derive general and replicable rules from existing examples. Consequently, it is challenging for a DCNN model to fully interpret complex cultural landscapes solely based on the visual elements depicted in images.

Perception of places is also influenced by subjective meaning. An individual's sense of place is shaped by their personal experiences, life stage, and individual tastes and preferences \citep{Tuan1977Experience}. For example, the Temple Mount in Jerusalem holds different religious significance for Jews, Muslims, and Christians. Moreover, a place can vary significantly depending on the time of day, day of the week, or time of year. The concept of "spatial ethnography" developed by \citet{kim2015sidewalk} reveals the diverse forms of ``time-sharing'' within socially, culturally, and economically complex sidewalk streetscapes. In summary, while places hold unique meanings for individuals and groups, deep learning techniques can only summarize and infer \emph{collective} knowledge, disregarding idiosyncratic yet significant factors. A DCNN model may effectively count various human activities along sidewalks but will fail to capture how people from different demographics perceive and utilize public space. Incorporating individual and group preferences into AI studies is crucial for enhancing the representativeness of future AI models.

\subsection{Dealing with uncertainty in street-level imagery}
Uncertainty is an inherent characteristic of spatial data, and street-level imagery is no exception. Several sources of uncertainty can affect street-level imagery, including the Modifiable Areal Unit Problem (MAUP), ecological fallacy, measurement uncertainty, and temporal uncertainty.

The MAUP refers to the statistical bias that arises when point-based measurements are aggregated into zones \citep{fotheringham1991modifiable}. Street-level imagery is not uniformly distributed spatially. Map service imagery is constrained by the road network, and social media photos vary in distribution based on urban functions and human activity intensity. Even along a long street, the characteristics of the imagery tend to exhibit internal homogeneity, failing to capture the distinct appearances of parallel streets. Aggregating a limited number of image samples into spatial units can lead to vastly different outcomes, exacerbating the MAUP.

Aggregation of street-level imagery can also lead to the ecological fallacy, which occurs when individual image-level conclusions are drawn based on their aggregation. For example, wrongly assuming that all locations on a street are beautiful solely because the average beauty score of that street is high.

Measurement uncertainty is another challenge in street-level imagery, primarily caused by variations in camera-to-scene distances. The camera's position influences the proportions of visual features captured in the image, subsequently impacting their analysis by computer vision algorithms. For instance, street-level imagery of tall structures may only represent the parts closest to the ground level.

Temporal changes can also introduce uncertainty in street-level imagery, with seasonal variations affecting vegetation, sky view, and pedestrian activity. Unfortunately, most current studies rely on infrequent data collection, often overlooking these aspects. Google Street View imagery, for instance, is typically updated annually, limiting the analysis of neighborhood-level changes. However, the growing availability of granular and frequent datasets such as point clouds (LiDAR data) and crowdsourced initiatives like Mapillary offers promising opportunities for comprehensive research on the physical transformations of neighborhoods.


\subsection{Promising avenues of inquiry and future work}

Street-level imagery and deep learning techniques go beyond efficient measurements; they have the potential to uncover new understandings and knowledge about cities. In this section, we discuss several aspects that can be explored in future work.

One avenue for gaining new knowledge about urban functioning is by studying hidden visual cues in images, such as written language or signs of social disorder. Street-level imagery can easily capture written language in street names, business names, and advertisements, enabling the mapping of points of interest and service locations like restaurants, pawnbrokers, and payday loan outlets. Street signage can also provide insights into linguistic or ethnic groups, shedding light on population composition, social disorder, psychosocial stress, and other important yet often overlooked aspects of neighborhoods. For instance, researchers like \citet{sampson2004seeing} have demonstrated that broken windows, graffiti, and litter—easily extractable from street-level imagery—serve as proxies for measuring social disorder in neighborhoods.


The analysis of vehicle types in street-level imagery can also contribute to understanding the social dimensions of a city. Seminal work by \citet{gebru2017using} utilized vehicle types mined from images to infer neighborhood demographics and political tendencies. Additional vehicle characteristics, such as type (commercial/private), spatiotemporal presence patterns (captured by cameras), and the cost of private vehicles, can further enhance our understanding of the socioeconomic attributes of neighborhoods. One can imagine other visual cues from street-level imagery serving as important indicators of neighborhood dynamics.



AI scene-generation techniques combined with design criteria offer the possibility of creating entirely new cities. The use of Generative Adversarial Networks (GANs) \citep{goodfellow2014generative}, a special architecture of deep convolutional neural networks (DCNNs), allows the generation of realistic urban scenes based on learned knowledge from real street scenes. GANs enable computer-generated urban scenes based on user inputs, such as objective characteristics extracted from images (e.g., buildings, roads, and vehicles) or encoded perceptions (e.g., attractiveness, safety, and liveliness of a place). Urban scenes can also be generated and modified based on specific attributes of the physical environment \citep{bau2020understanding, zhu2020domain,Richter_2021}. This approach has applications in scenario planning, urban design, and high-resolution imagery generation, providing participants with the ability to envision cities that do not yet exist \citep{wu2022ganmapper,zhao2021deep}. Notably, \citet{noyman2020deep} developed a physical platform that allows users to generate street scenes by combining a wide range of street elements based on their preferences, including land-use types, road types, building density, and the presence of sidewalks \footnote{https://www.media.mit.edu/projects/deep-image-of-the-city/}.

The design of interpretable and reliable AI models has garnered increasing attention. Interpretable models can support the analysis of the urban physical environment in two key ways. First, in scientific research, machine learning models surpass traditional regression models in terms of fitting and modeling capabilities, enabling better predictions of human activity patterns and socioeconomic profiles in cities. Second, for practitioners and policymakers, interpretable machine learning models can reveal confounding factors inherent in causal analysis, facilitating more informed policy formulation and implementation compared to relying solely on research findings that establish clear causal pathways.

Lastly, the fine-grained characteristics of the physical environment derived from street-level imagery offer tremendous opportunities to systematically understand the spatial laws governing urban spaces. We posit that recognizable patterns exist in the organization of physical features within cities, suggesting the existence of a fundamental spatial unit that constitutes urban space. Future research could investigate how the computational representation of physical features changes as spatial scales and organizational units vary, exploring whether a consistent spatial scale can effectively represent the physical urban environment.

\section{Conclusion}
\label{sec:conclusion}

Using visual information to understand cities has a rich history in urban studies, city planning, and design. However, assessing the physical environment and people's responses to it has posed challenges until recently. The advent of artificial intelligence now provides us with more efficient and effective tools to comprehend the city and its impact on its residents.
This paper undertakes a review and comparison of traditional and contemporary literature on the visual analysis of cities. We propose a conceptual framework, termed \textit{Urban Visual Intelligence}, to summarize and guide our discussion on how digital technology, particularly street-level imagery and visual intelligence techniques, is transforming our understanding of cities and opening up new research avenues.
Ultimately, these new tools enable us to reexamine the classic theories and themes that have shaped the comprehension and design of cities for over a century. They have the potential to assist cities in creating environments that align more closely with human aspirations and behaviors in the digital age.
 




\bibliography{ref_ceus.bib, ref_kang.bib,ref_yuan.bib}

\end{document}